\pdfoutput=1

\documentclass[11pt]{article}

\usepackage[final]{acl}

\usepackage{times}
\usepackage{latexsym}

\usepackage[T1]{fontenc}

\usepackage[utf8]{inputenc}
\usepackage{microtype}
\usepackage{inconsolata}
\usepackage{graphicx}
\usepackage{rotating}
\usepackage{amsmath}

\title{Do LLMs Give Psychometrically Plausible Responses \\ in Educational Assessments?}

\author{
  Andreas Säuberli\textsuperscript{1,2}\hspace{1cm}Diego Frassinelli\textsuperscript{1}\hspace{1cm}Barbara Plank\textsuperscript{1,2} \\
  \textsuperscript{1}MaiNLP, Center for Information and Language Processing, LMU Munich, Germany \\
  \textsuperscript{2}Munich Center for Machine Learning (MCML), Munich, Germany \\
  \texttt{\{andreas.saeuberli, diego.frassinelli, b.plank\}@lmu.de}
}

\begin{document}
\maketitle
\begin{abstract}
  Knowing how test takers answer items in educational assessments is essential for test development, to evaluate item quality, and to improve test validity. However, this process usually requires extensive pilot studies with human participants. If large language models (LLMs) exhibit human-like response behavior to test items, this could open up the possibility of using them as pilot participants to accelerate test development. In this paper, we evaluate the human-likeness or \emph{psychometric plausibility} of responses from 18 instruction-tuned LLMs with two publicly available datasets of multiple-choice test items across three subjects: reading, U.S.\ history, and economics. Our methodology builds on two theoretical frameworks from psychometrics which are commonly used in edu\-cational assessment, \emph{classical test theory} and \emph{item response theory}. The results show that while larger models are excessively confident, their response distributions can be more human-like when calibrated with temperature scaling. In addition, we find that LLMs tend to correlate better with humans in reading comprehension items compared to other subjects. However, the correlations are not very strong overall, indicating that LLMs should not be used for piloting educational assessments in a zero-shot setting.
\end{abstract}

\begin{figure}
  \centering
  \includegraphics[width=\columnwidth]{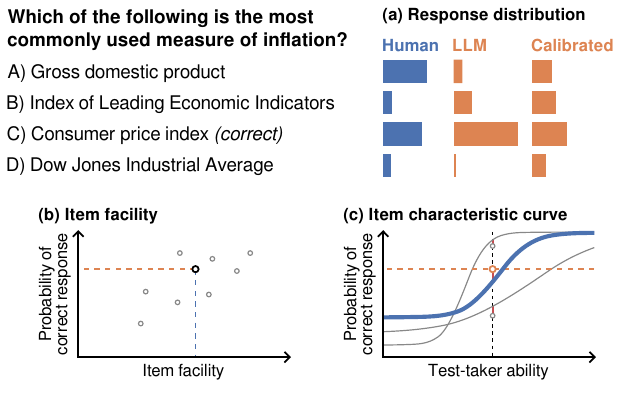}
  \caption{Example item from the NAEP dataset and illustration of our psychometric analyses of LLM responses. We use the first-token probabilities produced by LLMs and analyze how well they correspond to human test taker responses. Specifically, we look at \textbf{(a)} the similarity between LLM and human response distributions, \textbf{(b)} whether items that are difficult for humans are also difficult for LLMs, and \textbf{(c)} how well response probabilities in LLMs match those expected from humans.}
  \label{fig:example}
\end{figure}

\section{Introduction}
\label{sec:introduction}

Assessing students' knowledge and skills represents an important part of education: admission to universities, scholarship awards, and even political decisions on education policy are often based on large-scale educational assessments. Developing such high-stakes tests is a long and expensive process involving experts writing and reviewing test items and repeated piloting with hundreds or thousands of participants \citep{Green2020,Papageorgiou2021}. Therefore, the automation of parts of this process has been a long-standing topic in assessment research and practice \citep{Haladyna2013,Kurdi2019}. Most recently, large language models (LLMs) have been explored for tasks like item generation or item difficulty prediction \citep{Attali2022,Yaneva2024,Owan2023,May2025}.

The present work explores the possibility of using LLMs as participants of a pilot study in test development. A pilot study involves collecting and analyzing responses by human test takers to identify low-quality items and to measure item characteristics like difficulty. The statistical analysis of item responses most commonly follows one of two psychometric theories, classical test theory (CTT) or item response theory (IRT) \citep{Chang2021}. For LLMs to be useful models of human test takers, their responses must be human-like when analyzed within those theoretical frameworks -- we call this \textbf{psychometric plausibility}. This includes, for example, that items that are difficult for humans should also be difficult for LLMs. We propose an approach to evaluate the psychometric plausibility of LLM response distributions in multiple-choice test items, which is summarized in Figure~\ref{fig:example}.

Our contributions are two-fold: First, we present methods for assessing the psychometric plausibility of LLM responses with CTT and IRT (Section~\ref{sec:method}). Second, we benchmark the psychometric plausibility of 18 instruction-tuned LLMs across two datasets and three test subjects, showing that none of the models are sufficiently reliable to simulate test takers for piloting (Section~\ref{sec:experiment}).

\section{Related work}

A growing body of research has studied the use of natural language processing (NLP) for analyzing or evaluating test items. Examples of specific tasks are predicting difficulty \citep{Yaneva2024}, evaluating answerability or guessability \citep{Raina2023,Saeuberli2024}, evaluating the quality of generated items \citep{Raina2022,Gorgun2024}, or predicting correlations between items \citep{Hernandez2022}. Some of these studies used NLP models to simulate test takers: \citet{Lalor2019} and \citet{Byrd2022} used ``artificial crowds'', i.e., a large number of models trained on subsampled or partially corrupted data, to simulate test takers at different ability levels. More recently, LLMs have been used. For example, \citet{Lu2024} and \citet{Hayakawa2024} applied prompting techniques to simulate multiple test takers with a single LLM. \citet{Park2024} and \citet{LaverghettaJr2022} used multiple models to represent a group of test takers, while \citet{Liusie2023} and \citet{Zotos2025} used LLM uncertainty as a proxy for predicting student's response distributions.

Simulating test takers makes it easy to generate large numbers of item responses, which in turn makes statistical item analysis feasible. For example, \citet{Liusie2023} and \citet{Hayakawa2024} used CTT to compare item difficulty between humans and LLMs, while \citet{Lalor2019}, \citet{Byrd2022}, and \citet{Park2024} predicted IRT-based item characteristics. \citet{LaverghettaJr2022} compared both CTT- and IRT-based item difficulty between humans and models.

Apart from the application of educational assessment, the human-likeness of predicted response distributions has also been studied in the context of human label variation in tasks with inherent disagreement between annotators \citep{Plank2022}. Techniques like temperature scaling or fine-tuning on soft labels have been employed to align predictive probabilities with human response distributions \citep{Baan2022,Chen2024}.

Our approach combines ideas from several of these works. Our aim is to measure whether the response probabilities of a single model can be a plausible representative of a single test taker or a group of test takers. In this study, we use temperature scaling to optimize the response distributions, leaving other calibration methods as future work. We draw from both CTT and IRT for evaluation.

\section{Psychometric plausibility}
\label{sec:method}

Psychometrics is concerned with the measurement of unobserved latent variables based on observed responses to test items. Examples of possible latent variables include language proficiency, intelligence, and personality traits like introversion. In educational assessment, two theoretical frameworks are commonly applied: \textbf{classical test theory (CTT)} and \textbf{item response theory (IRT)}. These theories model the ability of test takers based on their observed test scores, but they also allow us to analyze characteristics of test items such as their difficulty or discriminating power \citep{Livingston2011}. For this reason, CTT and/or IRT is often used in pilot studies during test development in order to identify low-quality items and improve test reliability.

In our approach to evaluating psychometric plausibility, we focus on item analysis, i.e., determining item characteristics based on item responses by humans or LLMs. The key idea is that a psychometrically plausible LLM should give responses that are aligned with the characteristics of the items as measured using human responses.

In the following subsections, we introduce the relevant basics of CTT and IRT. We then describe how the response distributions of LLMs can be evaluated in the context of these two theories.

\subsection{Classical test theory}

CTT models assume that the observed test score achieved by a test taker is the sum of the true test score (reflecting the test taker's ability) and a random error score \citep{Hambleton1993}. Item analysis usually involves calculating two statistics for each item:

\begin{itemize}
  \item \textbf{Item facility} is the proportion of test takers who answered the item correctly. High item facility corresponds to low item difficulty.
  \item \textbf{Item discrimination} is the correlation between a person's score on the item and their score in the entire test. Low discrimination indicates that the item is inappropriate for measuring the latent variable and might need to be removed from the test.
\end{itemize}

\subsection{Item response theory}

IRT introduces a set of probabilistic models that predict the response of a specific person to a specific item, taking into account the person's latent variable (e.g., ability) and the item's characteristics (e.g., difficulty and guessability). The definition of the IRT model depends on the choice of item characteristics involved and the response variable type. Here we focus on the \textbf{three-parameter logistic (3PL) model} for dichotomous (correct/incorrect) responses:
\begin{equation}
  P(X_{p,i} = 1) = c_i + \frac{1 - c_i}{1 + e^{-a_i (\theta_p - b_i)}}
  \label{eq:3plm}
\end{equation}

$X_{p,i}$ equals $1$ if person $p$ answered item $i$ correctly and $0$ otherwise. $\theta_p$ is the ability parameter for person $p$, and $a_i$, $b_i$, and $c_i$ are item characteristic parameters for item $i$.

\begin{itemize}
  \item $a_i$ reflects \textbf{discrimination}, i.e., how good the item is at distinguishing between more and less proficient test takers, similar to the discrimination parameter in CTT.
  \item $b_i$ is the \textbf{difficulty} parameter and reflects the level of ability required for a substantial increase in correct response probability.
  \item $c_i$ is the \textbf{guessing} parameter and corresponds to the probability with which a person can answer the item correctly even if it is much too difficult for their ability level.
\end{itemize}

Once fitted on a large number of test taker responses, an item's parameters define the shape of its \textbf{item characteristic curve} (ICC; see Figure~\ref{fig:example} (c) for examples), and allow us to predict the probability of a correct response given their ability level.

One important advantage of IRT over CTT is that item characteristics are not dependent on the sample of test takers who answered this item. Even if not every person answered every item, the parameters can still be compared between items, since they are estimated in the context of person abilities. A disadvantage of IRT is that it generally requires larger sample sizes \citep{Hambleton1993,Fan1998}.

\subsection{Psychometric plausibility of LLM responses}

For a LLM to be considered psychometrically plausible, its response probabilities across different items should match the response patterns expected from humans. To evaluate this, we can use the item characteristics estimated from human responses using CTT or IRT.
In the following, we present two examples for such evaluations.

\paragraph{How well does a LLM fit CTT item facility statistics?}

To check this, we interpret the LLM's response probabilities as the response distribution in a sample of test takers. Specifically, the LLM should predict a higher probability for the correct answer on easier items compared to more difficult items. Therefore, we propose Pearson's correlation coefficient between human-based item facility and the LLM's probability for the correct response as an evaluation metric.

In the present paper, we focus on facility as the only CTT item statistic. Correlating with discrimination statistics would require response data at the level of individual test takers or pre-computed discrimination values, which are not available in the datasets we are using.

\paragraph{How well does a LLM fit IRT item characteristic curves?}

To evaluate this, we consider the LLM's response probabilities as representative of a single imaginary test taker with a specific ability. For example, the model may be calibrated to match the ability of an average test taker. Given each item's ICC, we can then compare the model's correct response probabilities to the ones predicted by the IRT model.

We will demonstrate these two analysis methods in the following experiment.

\section{Experimental setup}
\label{sec:experiment}

We empirically evaluate the psychometric plausibility of 18 LLMs across two datasets and three test subjects, comparing model and human response distributions and applying the analyses described in the previous section.

\subsection{Datasets}
\label{sec:experiment:datasets}

\paragraph{NAEP.}

The National Assessment of Educational Progress (NAEP) is a nation-wide and congressionally mandated educational assessment program in the United States.\footnote{\url{https://nces.ed.gov/nationsreportcard/about/}} NAEP involves tests across ten subjects at grades 4, 8, and 12. The tests include selected response items as well as constructed response items. A subset of items from previous years along with student response distributions and IRT item parameters are published and can be accessed online through the Questions Tool.\footnote{\url{https://www.nationsreportcard.gov/nqt/}} For our experiments, we used only four-option multiple-choice items from \emph{Reading}, \emph{U.S. History}, and \emph{Economics} tests, because most items in these subjects do not heavily rely on images, so that the LLM input can be text-only. For items that do include images, we included the alternative text and manually excluded items that were unanswerable without access to the full image. For some reading items, the full passage text was unavailable due to licensing issues -- we also excluded these items.\footnote{Refer to our code repository for detailed filter criteria and excluded items: \url{https://github.com/mainlp/llm-psychometrics}} This resulted in a total of 549 items, namely: 252 items in reading, 204 in history, and 93 in economics.

\paragraph{CMCQRD.}

The Cambridge Multiple-Choice Questions Reading Dataset (CMCQRD; \citealp{Mullooly2023}) contains four-option multiple-choice reading items for proficiency levels B1, B2, C1, and C2 in the Common European Framework of Reference for Languages (CEFR). Unlike NAEP, these items are targeted at L2 English learners. For a subset of the items, student response distributions and rescaled IRT difficulty parameters are provided. We included all items with available response distributions, resulting in a total of 504 items. Because the dataset's documentation does not include precise information about how the IRT parameters have been rescaled, it is impossible to reconstruct the original ICCs or interpret their meaning in relation to the test takers' abilities. Thus, we exclude the CMCQRD dataset from our IRT-based analysis.

\subsection{Language models}

We selected 18 recently published open-weight instruction-tuned LLMs\footnote{We also tested non-instruction-tuned LLMs. While the overall results are very similar, instruction-tuned models tended to slightly outperform base models. Therefore, we only report results from the instruction-tuned models here.} from four model families: Llama 3 \citep{Grattafiori2024}, OLMo 2 \citep{OLMo2025}, Phi 3/4 \citep{Abdin2024,Abdin2024a}, and Qwen 2.5 \citep{Qwen2024}. We included models ranging in size from 0.5B to 72B parameters to explore the effect of model capability on human-likeness of the responses. We used the implementations in the Hugging Face \emph{transformers} library \citep{Wolf2020}. Models with 70B or more parameters were loaded with 8-bit quantization.

\subsection{Prompting and response extraction}

We used a simple prompt with a user message instructing the model to select the correct answer option and to output only the corresponding letter (A, B, C, or D). The exact prompt template can be found in Appendix~\ref{sec:appendix-prompt}. We used the model's default system messages where applicable.

To get a probability distribution, we extracted the first predicted token logits for the four answer option letters and applied the softmax function. Since LLM responses are highly sensitive to the order of multiple-choice answer options \citep{Wang2024,Zheng2024,Pezeshkpour2024}, we prompted four times per item and reordered the options such that every option appears in every position exactly once, and averaged the probabilities from the four permutations. \citet{Zheng2024} showed that this ``cyclic permutation'' is practically as efficient for debiasing results as full permutation, which would require $4! = 24$ model passes.

\subsection{Temperature scaling}
\label{sec:experiment:temperature-scaling}

In preliminary experiments, we found that most LLMs (especially very large ones) tend to be overly confident compared to the human response distributions, assigning almost all probability mass to a single answer option. Temperature scaling is a common and effective approach to mitigate this issue and bring the uncertainty in LLM responses closer to human variability \citep{Guo2017,Baan2022,Chen2024}. It involves increasing the temperature parameter in the softmax calculation, essentially moving some probability mass from highly probable to less probable options.

\begin{figure*}
  \centering
  \includegraphics[width=\textwidth]{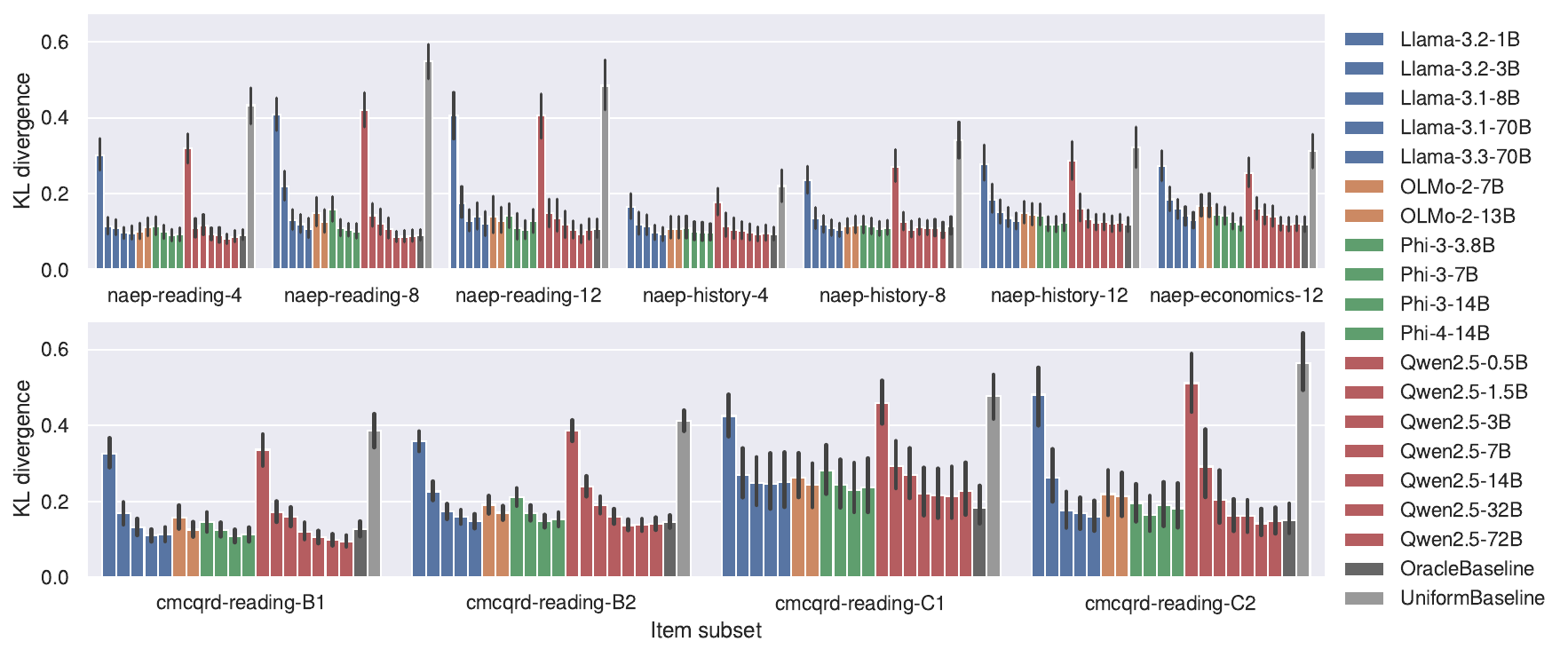}
  \caption{Mean KL divergence between temperature-scaled LLM response probability distributions and human response distributions. Models are colored by family and ordered by increasing number of parameters within families. Error bars are bootstrapped 95\% confidence intervals.}
  \label{fig:kldiv-optimized}
\end{figure*}

In our case, we find the optimal temperature that minimizes the Kullback-Leibler (KL) divergence between LLM and human response distributions (see Appendix~\ref{sec:appendix-temperature} for details). We apply this optimization separately to each LLM and each subset of items, i.e., each subject-grade combination in NAEP and each proficiency level in CMCQRD. This is important because the human response distributions are not sampled from the same population of test takers across all subsets (e.g., 4th grade items were only answered by 4th graders).

We perform the temperature optimization on the same data as the evaluation \citep[cf.][]{Baan2022,Liusie2023}. This means that the results should be considered an upper bound. In other words, we are testing the best-case scenario, where we have enough data to calibrate the LLMs perfectly to the human distributions as possible.

\subsection{Evaluation metrics}

We evaluate the human-likeness and psychometric plausibility of LLM responses from three perspectives:

Following \citet{Liusie2023} and \citet{Hayakawa2024}, we report the \textbf{average KL divergence} between the temperature scaled LLM and human response distributions. In addition to comparing the probability for the correct answer option, this metric also captures the similarity of the distractor probabilities.

For our \textbf{CTT-based analysis}, we report \textbf{Pearson's correlation coefficient} between the item facilities and the correct LLM response probabilities. This reflects the idea that psychometrically plausible LLMs should be more confident in the correct answer option when the item is easier.

In the \textbf{IRT-based analysis}, we assume that the temperature-scaled LLM response distributions reflect the response behavior of an average test taker, meaning a person with an ability parameter that is the mean of the sample. The ability parameters in NAEP's IRT models are fixed to have mean zero,\footnote{\url{https://nces.ed.gov/nationsreportcard/tdw/analysis/scaling_est.aspx}} therefore we use Equation \ref{eq:3plm} to calculate the expected correct response probability for human test takers with ability $\theta_p = 0$ for each item $i$:
\begin{equation}
  P_\text{expected}(X_i = 1) = c_i + \frac{1 - c_i}{1 + e^{a_i b_i}}
  \label{eq:3plm-theta0}
\end{equation}
We compare these values to the LLM's observed correct response probabilities and report \textbf{Pearson's correlation coefficient}.

\section{Results}

\subsection{Comparison of response distributions}

Figure~\ref{fig:kldiv-optimized} shows the average KL divergence between LLM and human response distributions, including two simple baselines: \textbf{UniformBaseline} always predicts the same probability (25\%) for all answer options. \textbf{OracleBaseline} always predicts the same probability for all distractors and a higher probability for the correct answer option (the same for all items). OracleBaseline is optimized using the same temperature scaling approach as the other models, as described in Section~\ref{sec:experiment:temperature-scaling}.

Across all model families and item subsets, we observe that LLM responses become more similar to the human distribution with increasing model size. However, only a small number of very large models in the CMCQRD B1 item subset managed to significantly outperform the OracleBaseline (bottom row in Figure~\ref{fig:kldiv-optimized}). This shows that the distribution of probabilities among distractors is not accurately modeled.

\subsection{CTT analysis}

\begin{figure}
  \centering
  \includegraphics[width=\columnwidth]{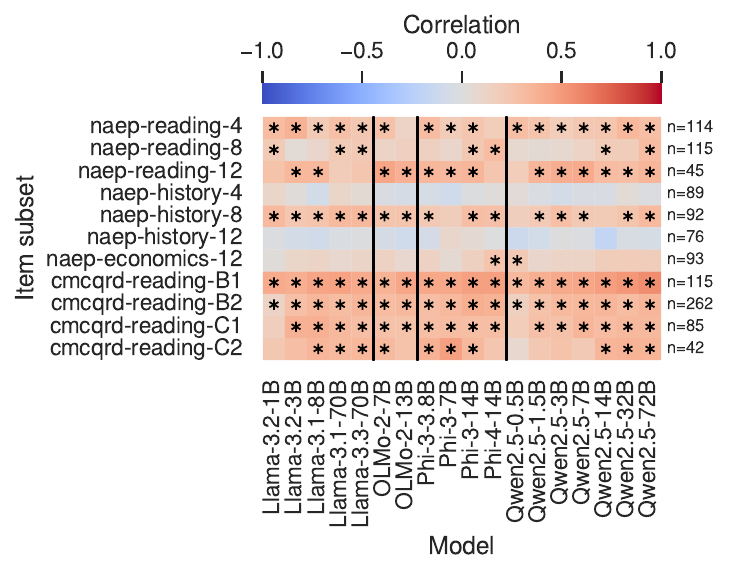}
  \caption{Pearson correlation between LLM correct response probabilities and item facilities. Numbers in the item subset labels refer to the grade level. $*$ denotes significance (two-tailed, $p < 0.05$), $n$ refers to the sample size in each cell of the corresponding row.}
  \label{fig:ctt-correlation}
\end{figure}

Correlations between the LLMs' correct answer probabilities and item facilities are visualized in Figure~\ref{fig:ctt-correlation}. While there does not seem to be a clear effect of model family or size, the correlations differ substantially between item subsets. The highest correlation coefficients were achieved in the CMCQRD B1 reading items, ranging from 0.32 to 0.56 across models. Among items from the NAEP datasets, most significant correlations can be found in reading items and 8th grade history items. However, the correlations are not strong overall and fluctuate substantially across grade levels.

\subsection{IRT analysis}

\begin{figure}
  \centering
  \includegraphics[width=\columnwidth]{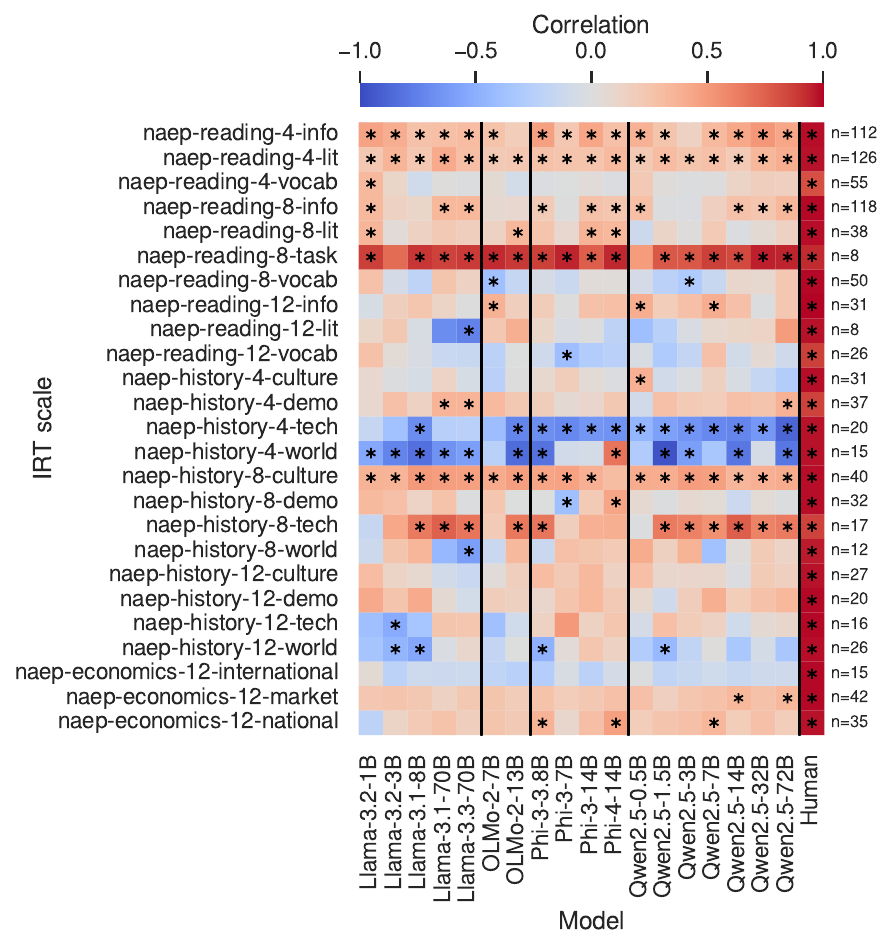}
  \caption{Pearson correlation between LLM correct response probabilities and expected correct response probabilities based on human IRT models. Numbers in the IRT scale labels refer to the grade level. $*$ denotes significance (two-tailed, $p < 0.05$), $n$ refers to the sample size in each cell of the corresponding row.}
  \label{fig:irt-correlation}
\end{figure}

NAEP considers multiple different skills for each subject (e.g., informational and literary reading skill) and therefore separate IRT models with different ability scales are fitted. Some items test multiple skills and are shared between different scales (but with different item parameters).

In Figure~\ref{fig:irt-correlation}, we report the correlations between LLM's correct answer probabilities and expected human correct response probabilities across NAEP IRT scales. As an upper bound, we also include the human response distributions as a model, i.e., the correlation between the response probability for the ``average'' test taker in the IRT model and the observed proportion of correct responses among human test takers (last column in Figure~\ref{fig:irt-correlation}).

Similar to the CTT results, most significant correlations can be found in reading items and 8th grade history items, and no effect of model family or size emerged. Notably, however, we also find significant \emph{negative} correlations in some 4th grade history items. This means that these LLMs tend to be \emph{more} confident in the correct answer when the item is more difficult, contradicting the expectations for psychometrically plausible responses.

Overall, while human correlations are consistently close to $1.0$, LLM correlations are rather low, and the number of significant correlations is small (considering that we expect 5\% of results to be type~I errors with the chosen significance level). However, given that the IRT analysis uses smaller item subsets and puts more stringent criteria on the LLM responses than the CTT analysis, these results are not overly surprising.

\section{Discussion}

The presented method is a multi-faceted approach, providing different perspectives on the human-likeness of LLM responses: The response distribution can tell us about a model's ability to model the success of distractors; the CCT analysis can show how well the model's probabilities represents a whole group of test-takers; and finally, the IRT analysis captures the plausibility of LLMs as an individual test taker in a specific skill.

\paragraph{LLMs are not easily distracted.}

Comparing the response distributions between humans and LLMs shows that especially large LLMs are good at predicting the \emph{correct} answer (see Appendix~\ref{sec:appendix-accuracy}), but bad at predicting which \emph{incorrect} answer options humans are likely to be distracted by (otherwise, they would outperform the OracleBaseline in Figure~\ref{fig:kldiv-optimized}). An example of this is also shown in Figure~\ref{fig:example}, where the item contains a very successful distractor (A), but the LLM (Qwen2.5-0.5B) assigns almost no probability mass to it. Calibration using temperature scaling cannot alleviate this issue, and reducing model size is not effective either (see Appendix \ref{sec:appendix-accuracy} for a more detailed analysis). This is an important limitation in applying LLMs for evaluating distractors.

\paragraph{Results are consistent across models, but inconsistent across subjects.}

While the correlations in the CTT and IRT analyses are likely too low to be useful for analyzing or evaluating single items, some interesting patterns can still be observed. The results are remarkably consistent across families and -- after calibration -- model sizes, demonstrating that all models are very similar to each other, but very dissimilar to humans in this setting (see Appendix~\ref{sec:appendix-ctt} for a more in-depth comparison).

At the same time, there are considerable differences between subjects and IRT scales. Correct answer probabilities appear to be more human-like in reading comprehension items compared to other subjects, while history items show mixed results, in some cases even eliciting strong negative correlations (see Figure~\ref{fig:irt-correlation}). This might indicate that reading comprehension in LLMs is more comparable to humans than other abilities such as long-term memory retrieval, which is required for answering test items in history and economics. Another possible explanation could be the fact that history and economics items more frequently contain images, which have to be understood from descriptions in the alternative text. Since we used text-only LLMs, this discrepancy in the way items were presented was inevitable. Future work could explore whether multimodal models are more successful with these item types.

\paragraph{How to improve psychometric plausibility?}

To a large degree, the lack of psychometric plausibility is in line with previous research \citep{Hayakawa2024,Zotos2025}. The success of attempts to make the model response distributions more human-like was very limited -- including our temperature scaling approach and \citeauthor{Hayakawa2024}'s \citeyearpar{Hayakawa2024} prompting techniques for injecting personas, uncertainty, or noise. Therefore, in order to improve psychometric plausibility, we will likely need to go beyond zero-shot prompting. Fine-tuning on human response distributions could be a promising direction for future research \citep[cf.][]{Chen2024}.

\section{Conclusion}

We demonstrated how LLM responses can be analyzed in the context of CTT and IRT and evaluated the human-likeness or psychometric plausibility of zero-shot responses. We found that neither reducing model size nor temperature scaling increased psychometric plausibility to a sufficient degree, but we observed slightly more human-like responses in reading comprehension compared to other subjects. We conclude that human-like response behavior in educational assessments has not emerged from the process of training instruction-tuned LLMs, calling for caution in their use. Fine-tuning on human response distributions may be necessary to create psychometrically plausible models that could be used for piloting.

\section*{Limitations}

\paragraph{Available item response data.}
Our analysis is limited by the type and amount of data available in the context of educational assessment. Item banks in high-stakes assessments are usually confidential to avoid leaking information for future test takers, and item responses from single test takers are generally not publicly released. Therefore, in order to keep our results reproducible, we only used publicly available datasets, where only aggregated response distributions and IRT parameters for a relatively small number of items are available. Given a larger amount of and less aggregated data, more fine-grained analyses would be possible (e.g., by including item discrimination in the CTT analysis) and more systematic patterns could be revealed.

\paragraph{Multimodal items.}
In addition, the NAEP dataset is not ideal for text-only LLMs, because some of the items involve extracting information from pictures. Although we replaced the pictures with alternative texts and manually removed unanswerable items (see Section~\ref{sec:experiment:datasets}), this could still have affected our results for this dataset.

\paragraph{Test-taker population.}
The two datasets we used contain response data from two different populations of test takers. While NAEP is targeted at children and adolescents (i.e., mostly L1 English speakers) in the U.S. school system, CMCQRD involves L2 learners of English. This difference could have affected the results and reduce the comparability between the two datasets.

\section*{Ethical considerations}

We see no ethical issues related to this work. All experiments were conducted with publicly available data and open-source software, and we have made all of our code openly available for reproducibility.\footnote{\url{https://github.com/mainlp/llm-psychometrics}} The two datasets we used only contain highly aggregated response data and do not include any information that could lead to the identification of individual test takers.

We used GitHub Copilot for coding assistance in the implementation of the experiment and the anlysis of the results. All generated code was manually checked and thoroughly tested.

\section*{Acknowledgements}

This research is in parts supported by the ERC Consolidator Grant DIALECT 101043235.

\bibliography{references}

\onecolumn
\appendix

\section{Prompt templates}
\label{sec:appendix-prompt}

The following prompt template was used for items with a reading passage (i.e., reading comprehension items):

\noindent\fbox{\parbox{\textwidth}{
    \tt
    Based on the following text, select the correct answer to the question below. \\\\
    Text: \textbf{\{passage\}} \\\\
    Question: \\
    \textbf{\{item stem\}} \\
    A) \textbf{\{option 1\}} \\
    B) \textbf{\{option 2\}} \\
    C) \textbf{\{option 3\}} \\
    D) \textbf{\{option 4\}} \\\\
    Respond only with the letter of the answer (A, B, C, or D).
}}

\noindent The following prompt template was used for items without a reading passage (i.e., history and economics items):

\noindent\fbox{\parbox{\textwidth}{
    \tt
    Select the correct answer to the following question. \\\\
    Question: \\
    \textbf{\{item stem\}} \\
    A) \textbf{\{option 1\}} \\
    B) \textbf{\{option 2\}} \\
    C) \textbf{\{option 3\}} \\
    D) \textbf{\{option 4\}} \\\\
    Respond only with the letter of the answer (A, B, C, or D).
}}

\section{Response accuracy}
\label{sec:appendix-accuracy}

\begin{figure*}[h]
  \centering
  \includegraphics[width=\textwidth]{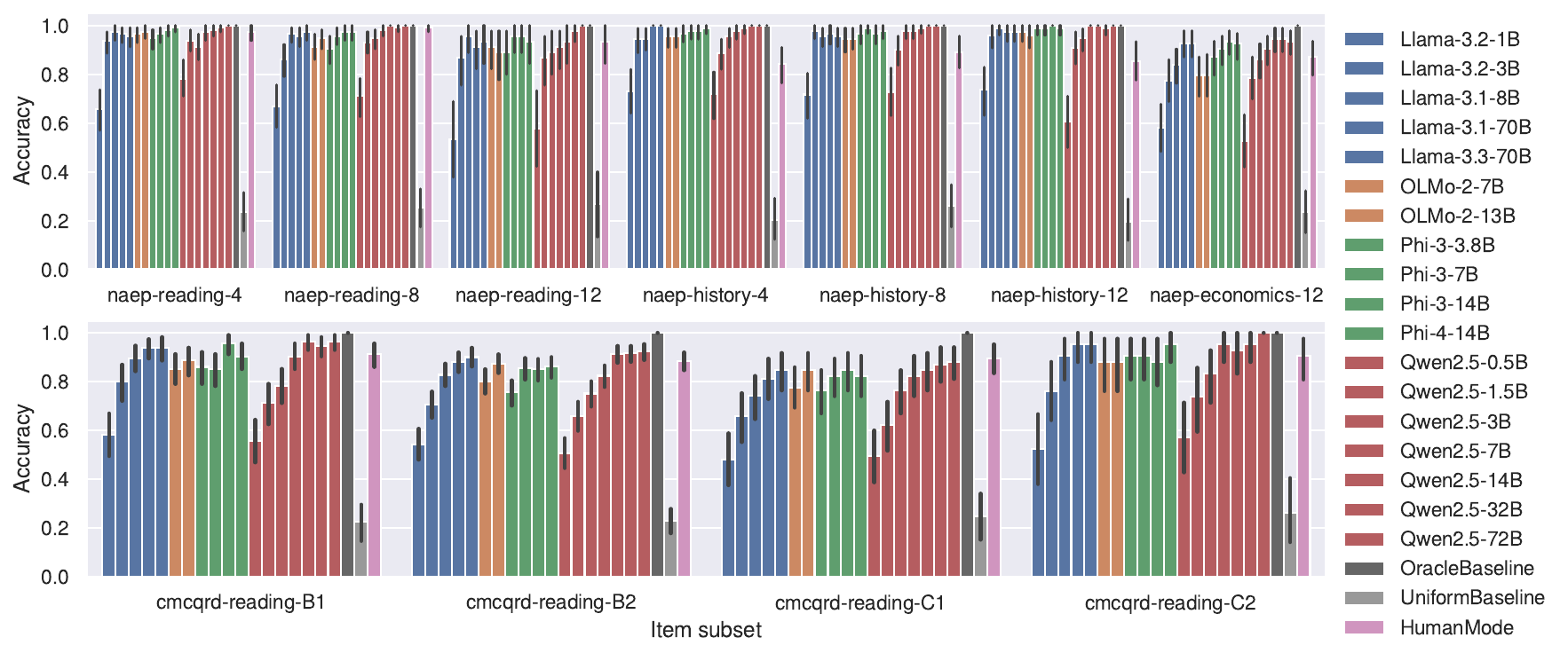}
  \caption{Mode accuracy across item subsets, models, baselines, and humans. Error bars are bootstrapped 95\% confidence intervals.}
  \label{fig:accuracy}
\end{figure*}

Figure~\ref{fig:accuracy} shows the mode accuracy of models and humans, i.e., the proportion of items where the option with the highest response probability is the correct one. The high accuracy of large models shows that the items are answerable given the available information in the prompt.

\section{Details on temperature scaling}
\label{sec:appendix-temperature}

\begin{figure*}[h]
  \centering
  \includegraphics[width=\textwidth]{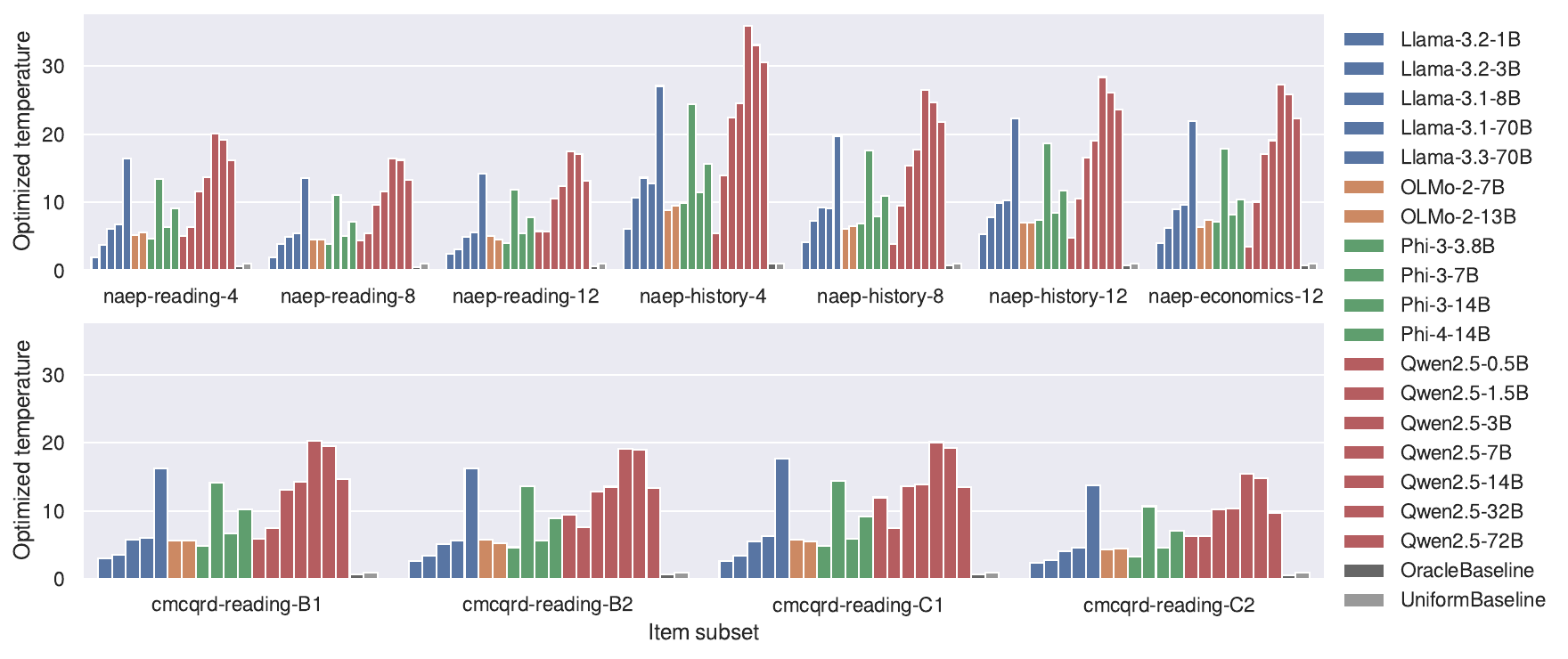}
  \caption{Optimized temperature value for each model and item subset.}
  \label{fig:temperature}
\end{figure*}

We optimized temperature parameters using KL divergence as a loss function and an Adam optimizer (see \texttt{analysis.py} in the code repository). The resulting optimized temperature values are visualized in Figure~\ref{fig:temperature}. Larger LLMs tend to be overly confident, assigning almost all probability mass to a single answer option, and therefore require higher temperatures to align them with human response distributions.

\begin{figure*}[h]
  \centering
  \includegraphics[width=\textwidth]{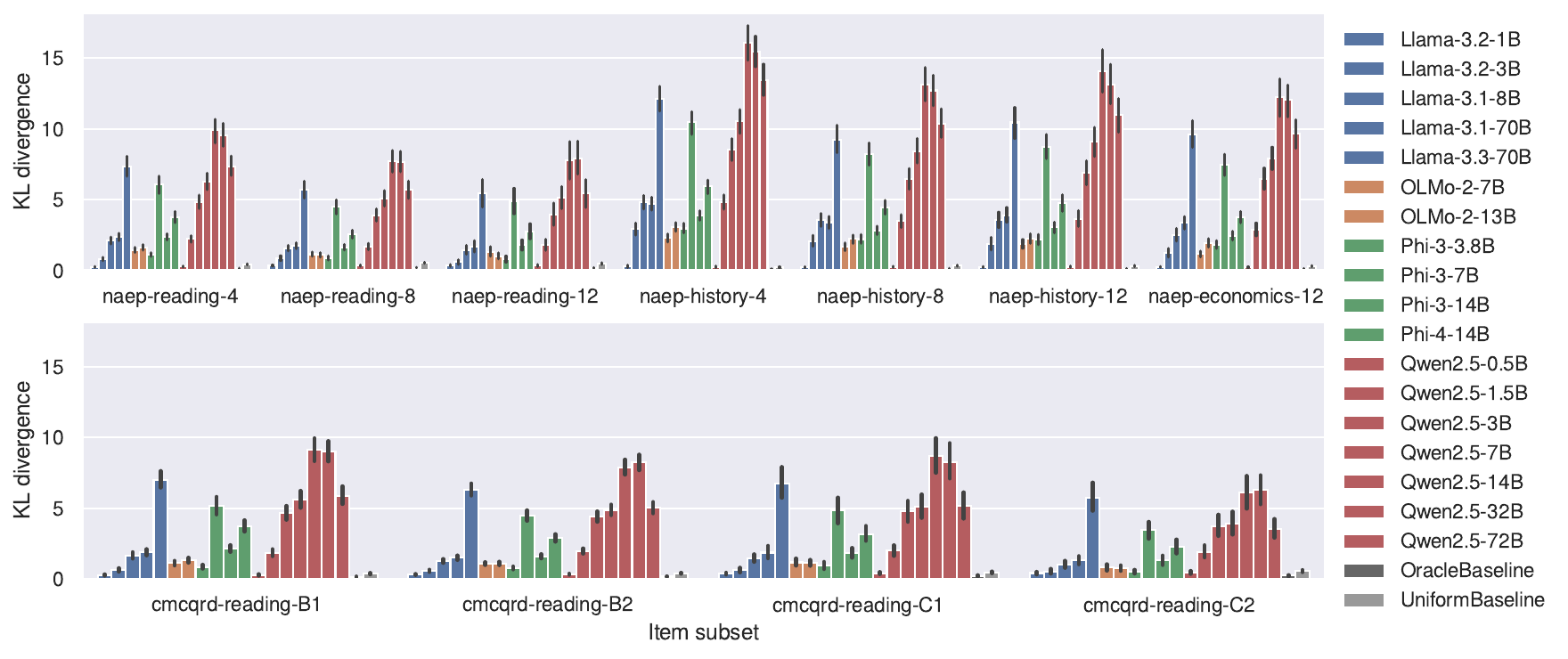}
  \caption{Mean KL divergence between LLM response probability distributions \emph{without} temperature scaling and human response distributions. Error bars are bootstrapped 95\% confidence intervals.}
  \label{fig:kldiv-unoptimized}
\end{figure*}

The effect of temperature scaling can be seen by comparing the results without temperature scaling in Figure~\ref{fig:kldiv-unoptimized} with Figure~\ref{fig:kldiv-optimized}.

\section{Additional results for CTT analysis}
\label{sec:appendix-ctt}

\begin{figure*}
  \centering
  \includegraphics[width=\textwidth]{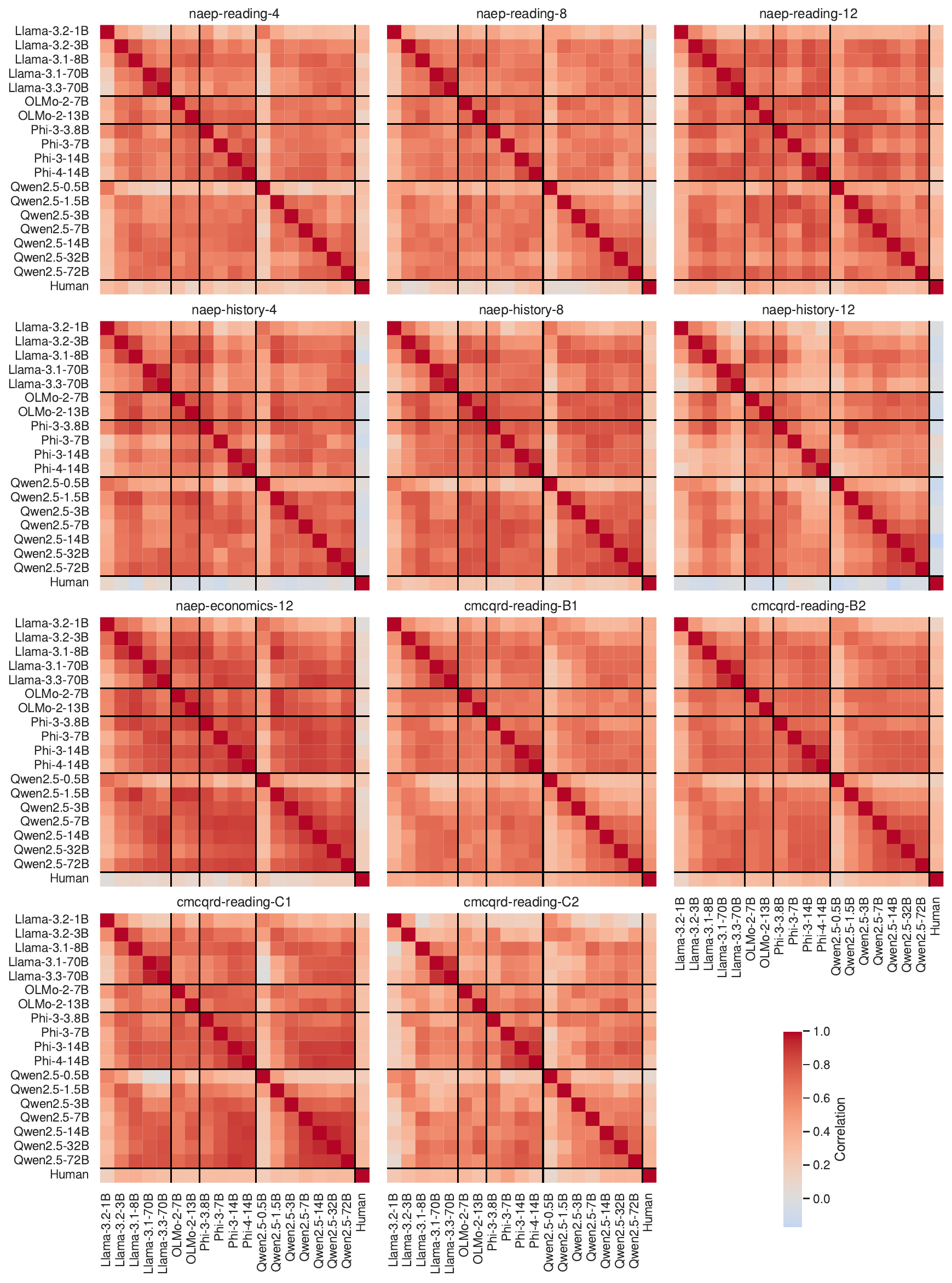}
  \caption{Pearson correlation between all LLM correct response probabilities and human item facilities.}
  \label{fig:ctt-correlation-full}
\end{figure*}

In addition to the correlations between models and humans in Figure~\ref{fig:ctt-correlation}, Figure~\ref{fig:ctt-correlation-full} shows the full correlation matrices, including model-model correlations. This confirms that the LLMs are much more similar to each other than to humans. In addition, models of similar sizes (but different model families) tend to be more similar to each other compared to models of different sizes.

\end{document}